\documentclass{article}

\usepackage{arxiv}

\usepackage[utf8]{inputenc} 
\usepackage[T1]{fontenc}    
\usepackage{hyperref}       
\usepackage{url}            
\usepackage{booktabs}       
\usepackage{amsfonts}       
\usepackage{nicefrac}       
\usepackage{microtype}      

\usepackage{booktabs} 
\usepackage{subfig}
\usepackage{graphicx,color}
\usepackage{colortbl}
\usepackage{algorithm}
\usepackage{algorithmic}
\usepackage{bm}
\usepackage{amsmath}
\usepackage{soul}
\usepackage{rotating}
\usepackage{multirow}
\usepackage{verbatim}

\usepackage{marvosym}
\usepackage{multiobjective}

\usepackage{ulem}
\usepackage{enumitem}
\setlist{leftmargin=4mm}

\usepackage[T1]{fontenc}

\usepackage{tabularx}

\newcounter{ALC@tempcntr}

\definecolor{mypink}{cmyk}{0, 0.7808, 0.4429, 0.1412}

\definecolor{blue-light}{RGB}{66, 191, 244}

\newcommand{\mar}[1]{\textcolor{magenta}{#1}}

\newcommand{\myr}[1]{\textcolor{mypink}{#1}}

\newcommand{\ignore}[1]{}

\title{On resampling vs. adjusting probabilistic graphical models in estimation of distribution algorithms}

\author{
Mohamed El Yafrani\\
Department of Materials and Production\\
Aalborg University\\
\texttt{mey@mp.aau.dk} \\
\AND
Marcella S. R. Martins\\
Federal University of Technology - Paran\'a (UTFPR)\\
\texttt{marcella@utfpr.com.br}\\
\AND
Myriam R. B. S. Delgado\\
Federal University of Technology - Paran\'a (UTFPR)\\
\texttt{myriamdelg@utfpr.edu.br}\\
\AND
Inkyung Sung\\
Department of Materials and Production\\
Aalborg University\\
\texttt{inkyung\_sung@mp.aau.dk}\\
\AND
Ricardo L\"uders\\
Federal University of Technology - Paran\'a (UTFPR)\\
\texttt{luders@utfpr.edu.br}\\
\AND
Markus Wagner\\
Optimisation and Logistics\\
The University of Adelaide\\
\texttt{markus.wagner@adelaide.edu.au}\\
}

\begin{document}

\maketitle

\begin{abstract}
The Bayesian Optimisation Algorithm (BOA) is an Estimation of Distribution Algorithm (EDA) that uses a Bayesian network as probabilistic graphical model (PGM). Determining the optimal Bayesian network structure given a solution sample is an NP-hard problem. This step should be completed at each iteration of BOA, resulting in a very time-consuming process. For this reason most implementations use greedy estimation algorithms such as K2. However, we show in this paper that significant changes in PGM structure do not occur so frequently, and can be particularly sparse at the end of evolution. A statistical study of BOA is thus presented to characterise a pattern of PGM adjustments that can be used as a guide to reduce the frequency of PGM updates during the evolutionary process. This is accomplished by proposing a new BOA-based optimisation approach (FBOA) whose PGM is not updated at each iteration. This new approach avoids the computational burden usually found in the standard BOA. The results compare the performances of both algorithms on an NK-landscape optimisation problem using the correlation between the ruggedness and the expected runtime over enumerated instances. The experiments show that FBOA presents competitive results while significantly saving computational time.
\end{abstract}


\section{Introduction}
\label{sect:sect1}

Probabilistic graphical models (PGMs) \cite{koller2009probabilistic} combine graph and probability theory to represent structured distributions. They play an important role in many computationally oriented fields \cite{Diakonikolas2016RobustLO}, including combinatorial optimisation, statistical physics, bioinformatics, machine learning, control theory and economics \cite{wainwright2008graphical}. PGMs are widely used in evolutionary optimisation, especially in Estimation of Distribution Algorithms (EDAs)~\cite{muhlenbein:96} when interactions among variables are considered (Multivariate EDAs).

EDAs are a class of Evolutionary Algorithms (EAs) that explores the search space by building a probabilistic model from a set with the current best candidate solutions~\cite{larranaga:02}. Since new solutions are sampled from the probabilistic model, the evolution is likely to be guided towards more promising areas of the search space.
Most of the Multivariate EDAs, such as Estimation of Bayesian network Algorithm (EBNA)~\cite{Etxeberria_and_Larranhaga:1999} and Bayesian Optimisation Algorithm (BOA)~\cite{Pelikan_et_al:1999}, use PGMs~\cite{Koller_and_Friedman:2009}, and in particular particular Bayesian networks (BNs), to be able to capture multivariate interactions between variables.

Finding the optimal structure of a Bayesian network is considered NP-hard \cite{heckerman:95}. Structure learning methods have been extensively studied in \cite{cooper:92, tsamardinos2003algorithms,Tsamardinos:06, aliferis2010local,abbeel2006learning,de2010combining, smith2010robustness, anandkumar2012learning,santhanam2012information,loh2012structure,bresler2014structure,bresler2015efficiently, Diakonikolas2016RobustLO}, resulting in a range of algorithms for various settings. However, constructing Bayesian network models is still particularly computationally expensive with the increasing number of variables and interactions.

In this work, we consider the Bayesian Optimisation Algorithm (BOA)~\cite{Pelikan_et_al:1999} for which we show experimentally that the changes in the PGM follow a certain pattern. This pattern is then used to propose an alternative version of BOA, called Fast BOA (FBOA) that decides whether to adjust\footnote{Adjusting a PGM here means generating a new PGM, either based on the previous one or not. } 
the PGM following a probability distribution. 

To achieve our goals, we start by analysing the behaviour of BOA and the similarities between consecutive Bayesian networks. Additionally, we explore the impact of problem features on the search performance of BOA versions while proposing a new runtime estimation metric which considers the computational complexity of evaluation and PGM generation. We consider the NK-model as our benchmark problem for this study. The results indicate that our model-based BOA is about three times faster than the standard BOA while producing quality solutions. 

This paper is organised as follows: 
Section~\ref{sec:sect2} provides the mathematical formulation of the NK-landscape model and briefly introduces Bayesian networks as PGM in EDAs.
Section~\ref{sec:sect3} presents the BOA framework used in our investigation and introduces a new runtime estimation approach. 
In Section~\ref{sec:sect4}, our model-based variant of BOA (FBOA) is introduced.
Results from numerical experiments are reported and discussed in Section~\ref{sec:sect5}.
Finally, Section~\ref{sec:conclusion} presents some conclusions and future directions. 

\section{Preliminaries}
\label{sec:sect2}

In this section, we first cover the basics of the family of problems with tunable ruggedness that we consider. Then, we describe the Bayesian networks and their use as probabilistic graphical models.

\subsection{NK-Landscape Model}

The NK-landscape models are a family of combinatorial problems proposed in~\cite{kauffman1993origins}
aiming to explore the way in which the neighborhood structure and the strength of the interactions
between neighboring variables (subfunctions) are linked to the search space ruggedness.

Let $ {\bf{X}}=(X_1,\ldots ,X_N)$ denote a vector of discrete variables and ${\bf{x}}=(x_1,\ldots ,x_N)$
an assignment to the variables.

An NK fitness landscape is defined by the following components:

\begin{itemize}
  \item Number of variables, $N$.
  \item Number of neighbors per variable, $K$.
  \item A set of neighbors, $\Pi(X_q) \in {\bf{X}}$, for $X_q$, $q \in \{1,\dots, N\}$ where $\Pi(X_q)$ contains $K$ neighbors.
  \item A subfunction $f_q$ defining a real value for each combination of values of $X_q$ and $\Pi(X_q)$, $q \in \{1,\dots, N\}$.
\end{itemize}

Both the subfunction $f_q$ for each variable $X_q$ and the neighborhood structure $\Pi(X_q)$
are randomly set.

The function $z_{NK}$ to be maximized is defined as:
\begin{equation}
  z_{NK}({\bf{x}}) = \sum_{q=1}^{N} f_q(x_q ,\Pi(x_q)). \label{eq:FNK}
\end{equation}

For a set of given parameters, the problem consists of finding the global maximum of the function $z_{NK}$~\cite{santana2015evolving}.

\subsection{Bayesian networks as PGM}

A Bayesian network is a mathematical structure developed to represent a joint probability distribution considering a set of variables. Bayesian networks are among the most general probabilistic models for discrete variables used in EDAs~\cite{larranaga:02,Koller_and_Friedman:2009}. In this paper, we use Bayesian networks to model multinomial data with discrete variables, generating new solutions using the particular conditional probability~\cite{Henrion86} described by Equation~\ref{eq:disc_dist}:

\begin{equation}
\label{eq:disc_dist}
p(y_m^k|\mathbf{pa}^{j,B}_m)=\theta_{y_m^k|\mathbf{pa}^{j,B}_m}=\theta_{mjk}
\end{equation}
where $\mathbf{Y}=(Y_1,...,Y_M)$ is a vector representation of $M$ random variables and $y_m$ its $m$-th component;  $B$ is the structure  and $\Theta$ a set of local parameters; $\mathbf{Pa}^B_m$ represents the set of parents of $Y_m$, which $\mathbf{pa}_m^{j,B} \in \{ \mathbf{pa}_m^{1,B},...,\mathbf{pa}_m^{t_m,B}\}$  denoting a particular combination of values for $\mathbf{Pa}^B_m$, $t_m$ is the total number of different possible instantiations of the parent variables of $Y_m$ given by $t_m=\prod_{Y_v\in \mathbf{Pa}_m^B} s_v$, with $s_v$ defining the total of possible values (states) that $Y_v$ can assume. The parameter $\theta_{mjk}$ represents the conditional probability that variable $Y_m$ takes its $k-$th value ($y_m^k$), knowing that its parent variables have taken their $j$-th combination of values ($\mathbf{pa}^{j,B}_m$). This way, the parameter set is given by $\Theta =\{\pmb{\theta}_1,..., \pmb{\theta}_m,... \pmb{\theta}_M \}$, where $\pmb{\theta}_m = (\theta_{m11},..., \theta_{mjk},..., \theta_{m,t_m,s_m} )$ and $M$ is the total number of nodes in the Bayesian network.
 
The parameters of $\Theta$ and $B$ are usually unknown, and the literature presents two possibilities to estimate them: Maximum Likelihood Estimate (MLE) and the more general Bayesian Estimate~\cite{bengoetxea:02}. In this work, we address the last method.

In terms of Bayesian network structures learning process, there are mainly three different approaches: score-based learning, constraint-based learning, and hybrid methods~\cite{Yuan:2013:LOB:2591248.2591250}. Score-based techniques apply heuristic optimisation methods to sort the structures selecting the one that maximizes the value of a scoring metric, like the K2 metric~\cite{cooper:92} in Equation~\ref{eq:k2}:

\begin{equation}
\label{eq:k2}
p(B|P)=\prod_{m=1}^{M}\prod_{j=1}^{t_m}\frac{(s_m-1)!}{(N_{mj}+s_m-1)!}\prod_{k=1}^{s_m}(N_{mjk})!
\end{equation}
where $N_{mjk}$ is the number of observations in the data set $P$ for which $Y_m$ assumes the $k$-th value given the $j$-th combination of values from its parents. 

Constraint-based learning methods typically use statistical tests to identify conditional independence relations from data and build a Bayesian network structure that best fits those relations~\cite{pearl:88}, such as Incremental Association Markov Blanket (IAMB)~\cite{tsamardinos2003algorithms} and Semi-Interleaved Hiton-PC (SI-HITON-PC)~\cite{aliferis2010local}. Hybrid methods aim to integrate the two approaches, like Max-Min Hill Climbing (MMHC)~\cite{Tsamardinos:06}.

In this paper we consider the K2 algorithm, which is a commonly-used, score-based greedy local search technique that applies the K2 metric (Equation~\ref{eq:k2}). It starts by assuming that a node, in a (pre-defined) ordered list, does not have any parent, then at each step it gradually adds the edges that increase the scoring metric the most, until no edge increases the metric anymore. 

\section{Revisiting Bayesian optimisation algorithms}
\label{sec:sect3}

In this section, we provide basic concepts used throughout this article (Sub-section \ref{sec:BOAframework}) and also present some contributions (Sub-sections \ref{sec:PGMadjust} and \ref{sec:ERT}) that will support the analysis performed later on. 


Among the most general probabilistic models for discrete variables used in EDAs are Bayesian networks. Several EDAs have been proposed based on this model, such as, EBNA and EBNAK2~\cite{Etxeberria_and_Larranhaga:1999}, BOA~\cite{Pelikan_et_al:1999} and hBOA~\cite{pelikan2008analysis}.

In this paper, we adopted BOA~\cite{Pelikan_et_al:1999} whose framework encompasses general steps and assumptions discussed in the next section.

\subsection{The Framework}
\label{sec:BOAframework}


The general framework for the EDA considered here is presented in Algorithm~\ref{alg:boa_framework}. The algorithm starts by generating a random initial Bayesian network $M^{(0)}$ (Step \ref{step:generate-pgm}). At each iteration, a set $S$ of $\mu$ solutions is generated in Step 4 
by sampling  solutions from the current model; and evaluated in Step 5. 
Then, the $\lambda$ best solutions in $S$ are stored in the current population $P$. Afterwards, the Bayesian network is adjusted based on the current population in Step~\ref{step:adjustment}. This process is repeated until a specified stopping criterion is met (e.g. maximum number of iterations, maximum fitness reached, etc.).

\begin{algorithm}[h]
\small \caption{General EDA/BOA framework} \label{alg:boa_framework}
\begin{algorithmic}[1]
\STATE $t \gets 0$
\STATE $M^{(0)} \gets initial\_model($\;$)$ \label{step:generate-pgm}
\REPEAT
  \STATE $S \gets sample\_solutions(M^{(t)})$ \label{step:sampling}
  \STATE $F \gets evaluate(S)$ \label{step:evaluation}
  \STATE $P \gets best\_solutions(S,F)$ \label{step:selection}
  \STATE $M^{(t+1)} \gets adjust\_model(P)$ \label{step:adjustment}
  \STATE $t \gets t+1$
\UNTIL{termination criterion met}
\end{algorithmic}
\end{algorithm}

\subsection{Tracing PGM adjustment}
\label{sec:PGMadjust} 

In a standard BOA implementation (Algorithm~\ref{alg:boa_framework}), the PGM is updated in each iteration. If two or more consecutive PGMs are very similar, then the algorithm performs a pointless time consuming task.

This is particularly costly for the problem addressed in this paper, as it is done many times for the multiple ruggedness levels and the multiple landscapes of the NK-models, besides it can be an issue for most of PGM-based approaches.

Aiming to quantify similarities between consecutive PGMs, we use in this work the Structural Hamming Distance ({\emph{SHD}})~\cite{Tsamardinos:06} as a metric. At each iteration of BOA, we store the {\emph{SHD}} value between two consecutive PGM structures, the current and the previous one, generating, at the end of evolution, an {\emph{SHD}} vector as large as the total number of iterations that were necessary to converge. 

In order to extract a common pattern across all multiple runs ($R$) of the algorithm, we need to aggregate these {\emph{SHD}} vectors. 
As the resulting {\emph{SHD}} vectors can be of different sizes due to the convergence speed of each run, we use a methodology to normalise the vector size, which is described in the following steps:

\begin{enumerate}
  \item Define the normalised size $L$ as the maximum size among all the {\emph{SHD}} vectors (i.e. the maximum number of iterations before convergence).
  \item For all the {\emph{SHD}} vectors with size smaller than $L$, fill-in the gaps by adding a '*' character in a uniform manner.
  \item To obtain the final aggregated normalised vector, joint all {\emph{SHD}} vector obtaining a matrix of size  $R \times L$. Then we average each column, considering that if the value corresponds to a '*', it is simply ignored. Afterwards, we normalise the vector elements into the range $[0,1]$.
\end{enumerate}

Then, at the end we have the aggregated normalised information stored in a vector named $\overline{SHD}$, which describes in average the pattern of PGM adjustments followed by BOA. 
We believe that this approach is a good fit to aggregate vectors in this particular situation compared to interpolation for instance. The reason we did not opt for a curve fitting approach was to avoid modifying the original  {\emph{SHD}} values and including additional values we are not sure are adequate.

\subsection{Estimating the runtime}
\label{sec:ERT}

In many studies on fitness landscape analysis~\cite{liefooghe2015feature, mar_cec:18, daolio2015global}, 
the number of fitness evaluations is used to estimate the runtime of a given stochastic algorithm.

Let $p_s \in (0; 1]$ be the probability of success of the algorithm and let $T_f$ be the random variable measuring the number of function evaluations for unsuccessful runs.

After $(t -1)$ failures, each one requiring $T_f$ evaluations, with the final successful run of $T_s$ evaluations, the total runtime is then defined as $T=\sum_{i=1}^{t-1}T_f+T_s$, where $t$ is the random variable measuring the number of runs. 
The random variable $t$ follows a geometric distribution with parameter $p_s$. By taking the expectation and by considering independent runs for each instance, stopping at the first success, we have:
\begin{equation}
\mathbf{E}[T]=(\mathbf{E}[t]-1)\mathbf{E}[T_f]+\mathbf{E}[T_s]
\end{equation}

The expected runtime for successful runs $\mathbf{E}[T_s]$ is estimated as the average number of function evaluations performed by successful runs, and we note $T_{max}$ the expected runtime for unsuccessful runs.
As the expectation of a geometric distribution for $t$ with parameter $p_s$ is equal to $1/p_s$, the estimated runtime can be expressed as the following:
\begin{equation}
\mathbf{E}[T]=\frac{1-\hat{p}_s}{\hat{p}_s}T_{max}+\frac{1}{t_s}\sum_{i=1}^{t_s}T_i
\end{equation}
where $t_s$ is the number of successful runs, $T_i$ is the number of evaluations for successful run $i$.

In the context of BOA, it is clear that this approach is far from being accurate. 
In fact, the process of adjusting the PGM has a higher computational complexity \footnote{In terms of complexity theory, finding the optimal PGM structure is NP-complete. However, some real-world problems could require more wallclock time for calculating the objective value of a single solution, in which case surrogate models are often used.}  than that of 
evaluating a population of solutions, even for fast greedy algorithms such as K2.
Therefore, we propose an alternative approach specific to our case study based on the expected time complexity instead of the number of evaluations. More precisely, we take into consideration the complexity of 
both: the cost of the objective function and the cost of generating a new PGM using a given structure learning algorithm like K2.

Let us start by estimating the time complexity of PGM adjustments of a BOA. We assume that it generates a Bayesian network based on a given probability distribution $\{p_j\}$. 
Let $U_s$ and $U_f$ denote the number of PGM adjustment for successful runs and unsuccessful runs respectively. 
We can define the number of PGM adjustment as $U=\sum_{i=1}^{t-1}U_f+U_s$. Thus, the expected number of PGM adjustment is defined by:
\begin{align}
\mathbf{E}[U] & = \frac{1-\hat{p}_s}{\hat{p}_s} U_f + U_s \nonumber \\
& = \frac{1-\hat{p}_s}{\hat{p}_s} \sum_{j=1}^{I_{max}} p_j + \frac{1}{t_s} \sum_{i=1}^{t_s} \sum_{j=1}^{I_i} p_j
\end{align}

By noticing that the complexity of evaluating a population of $\mu$  solutions, each one represented by an $N$ size vector,  is $\mu N$ and the complexity of the K2 algorithm is $2(N^5+N^4)$, and by unifying the runtime unit to \textit{number of elementary operation} instead of \textit{number of evaluation}, we obtain the following runtime equation:
\begin{equation}
ert = \mu N\mathbf{E}[T] + 2(N^5+N^4)\mathbf{E}[U]
\label{eq:ert}
\end{equation}
in our case $\mu=|S|$ is the sampled population size and $N$ the bitstring length.

\section{Designing a model-based BOA}
\label{sec:sect4}


In this section, we start by analysing the similarity patterns between PGM based on the output of a standard BOA implementation. The outcome is then used to propose a model-based algorithm based on the BOA framework that adopts a strategy to reduce the frequency of PGM adjustments.

\subsection{Analysis of PGM adjustment patterns}
In Figure \ref{fig:shd-fitness}, we show the evolution of  {\emph{SHD}} values (solid lines) and objective values (dashes lines) for 5 randomly selected runs (each color is associated with a specific run), for each ruggedness level $K$. 
The experiments are conducted for different problem sizes ($N=18$, $K=\{2, 4, 6, 8, 10, 12, 14, 16\}$) and the {\emph{SHD}} curves (each one with a different convergence speed meaning different number of iterations until find the best solutions) have been fitted with polynomials of degree $6$.

\begin{figure*}[h]
\centering
\subfloat[K=2]{
  \centering
  \includegraphics[scale=0.26]{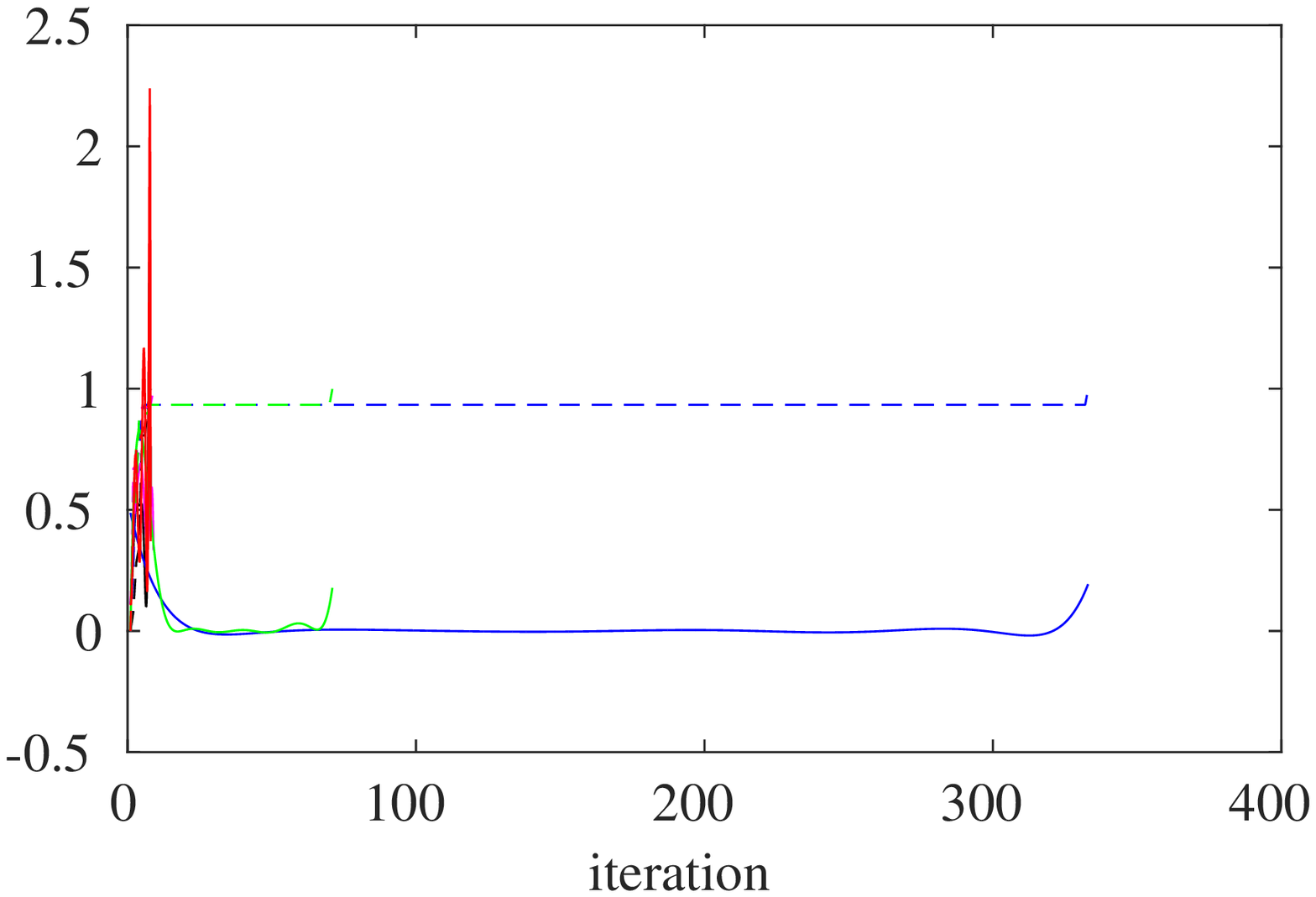}
  \label{fig:K2land5}
}
\quad 
\subfloat[K=4]{
  \centering
  \includegraphics[scale=0.26]{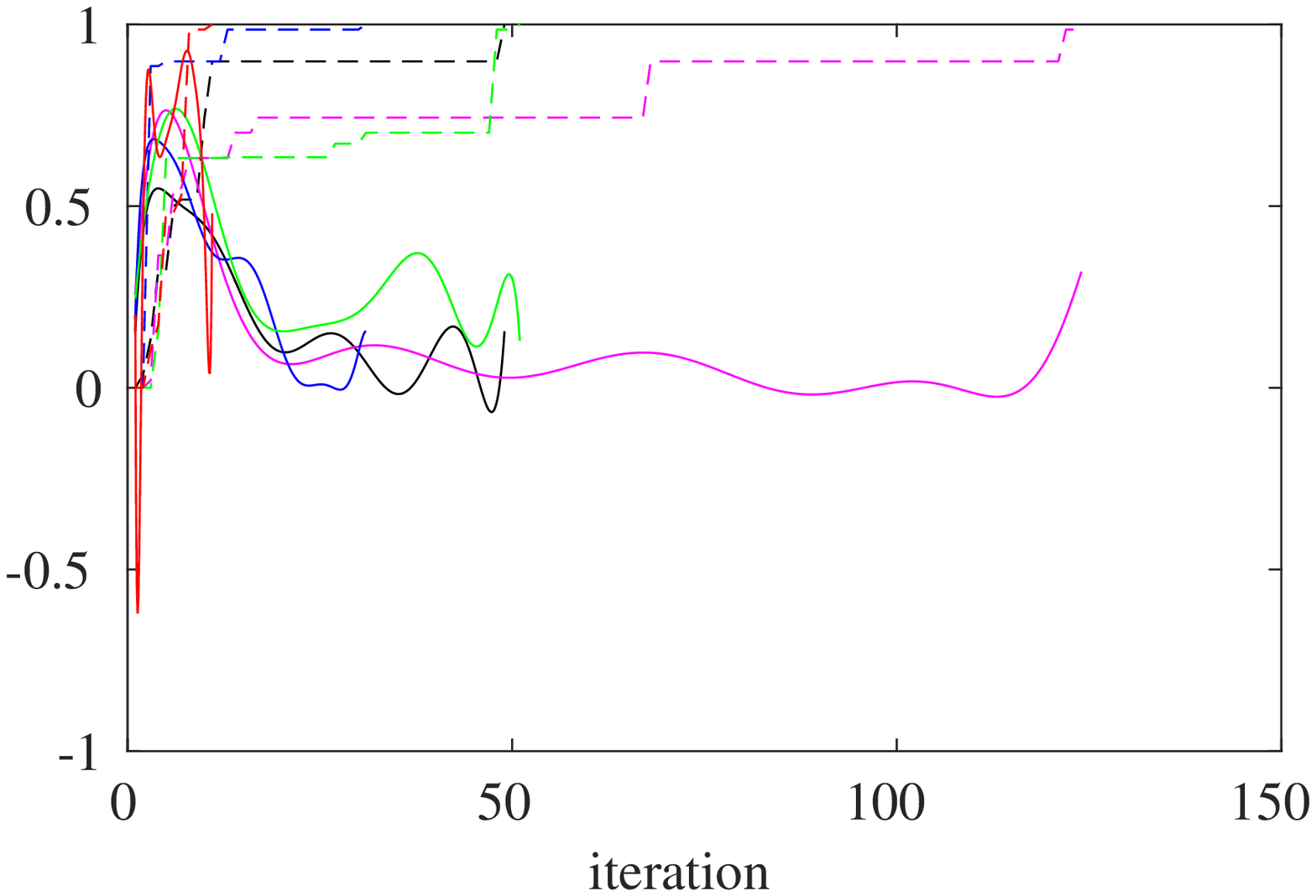}
  \label{fig:K4land5}
}
\quad 
\subfloat[K=6]{
  \centering
  \includegraphics[scale=0.26]{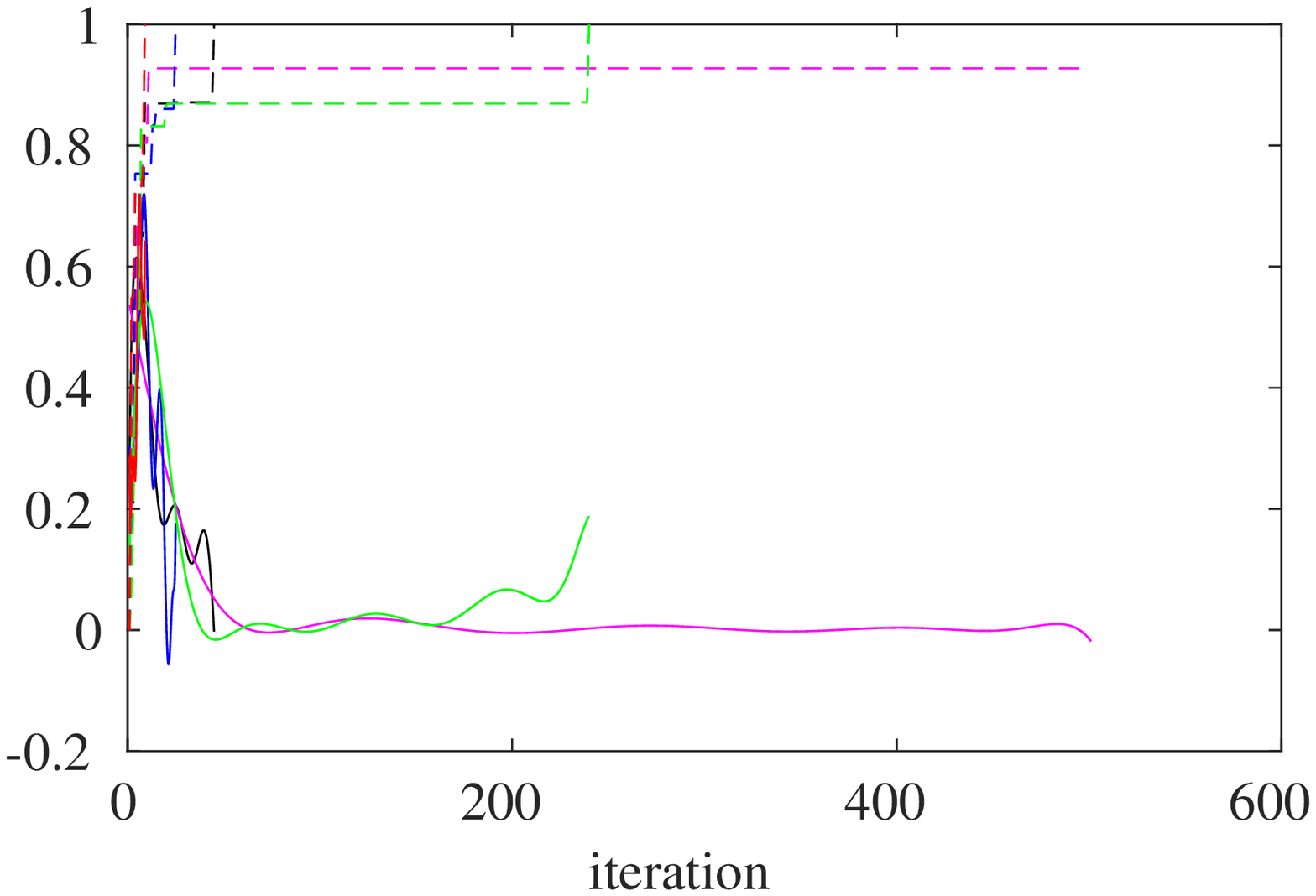}
  \label{fig:K6land5}
}
\quad 
\subfloat[K=8]{
  \centering
  \includegraphics[scale=0.26]{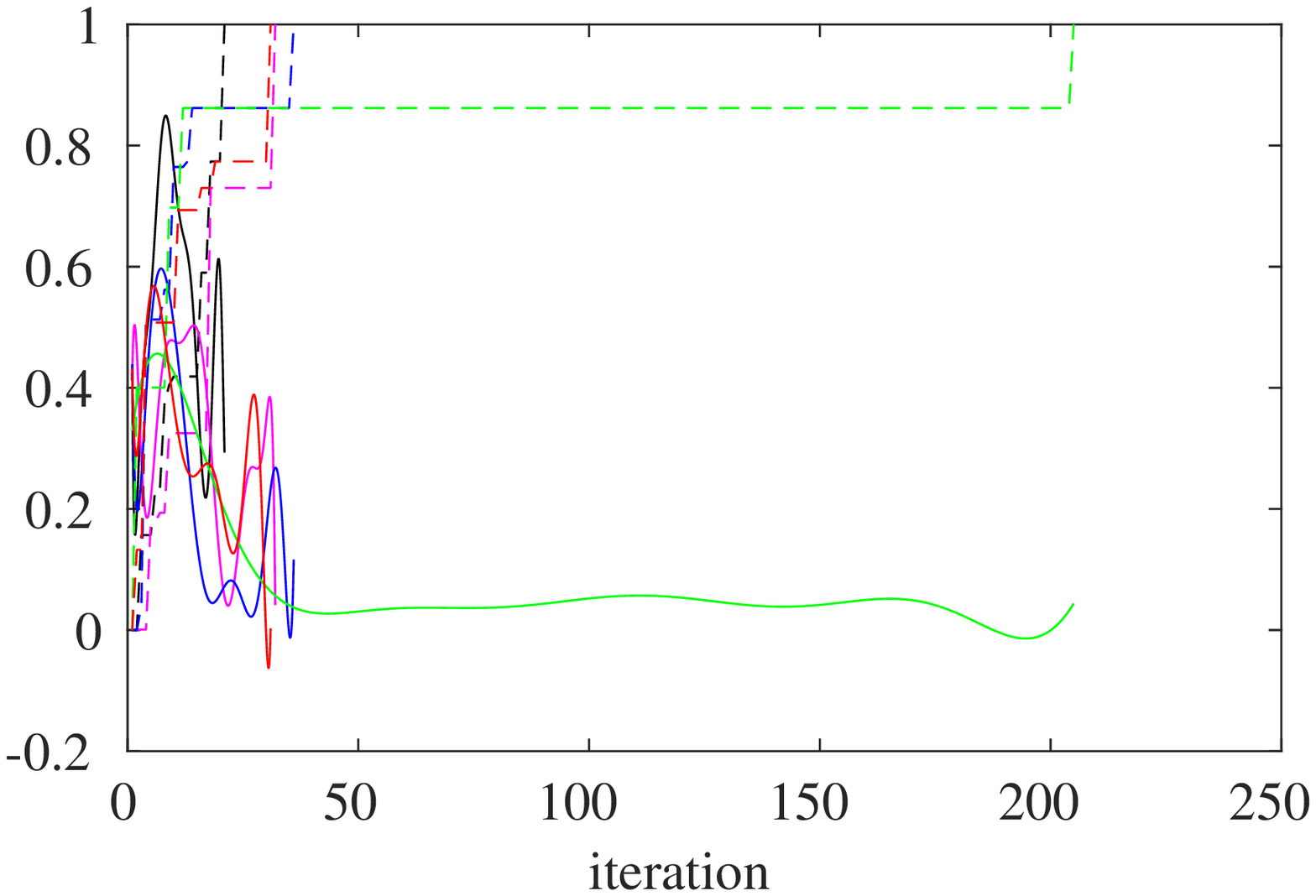}
  \label{fig:K8land5}
}
\quad 
\subfloat[K=10]{
  \centering
  \includegraphics[scale=0.26]{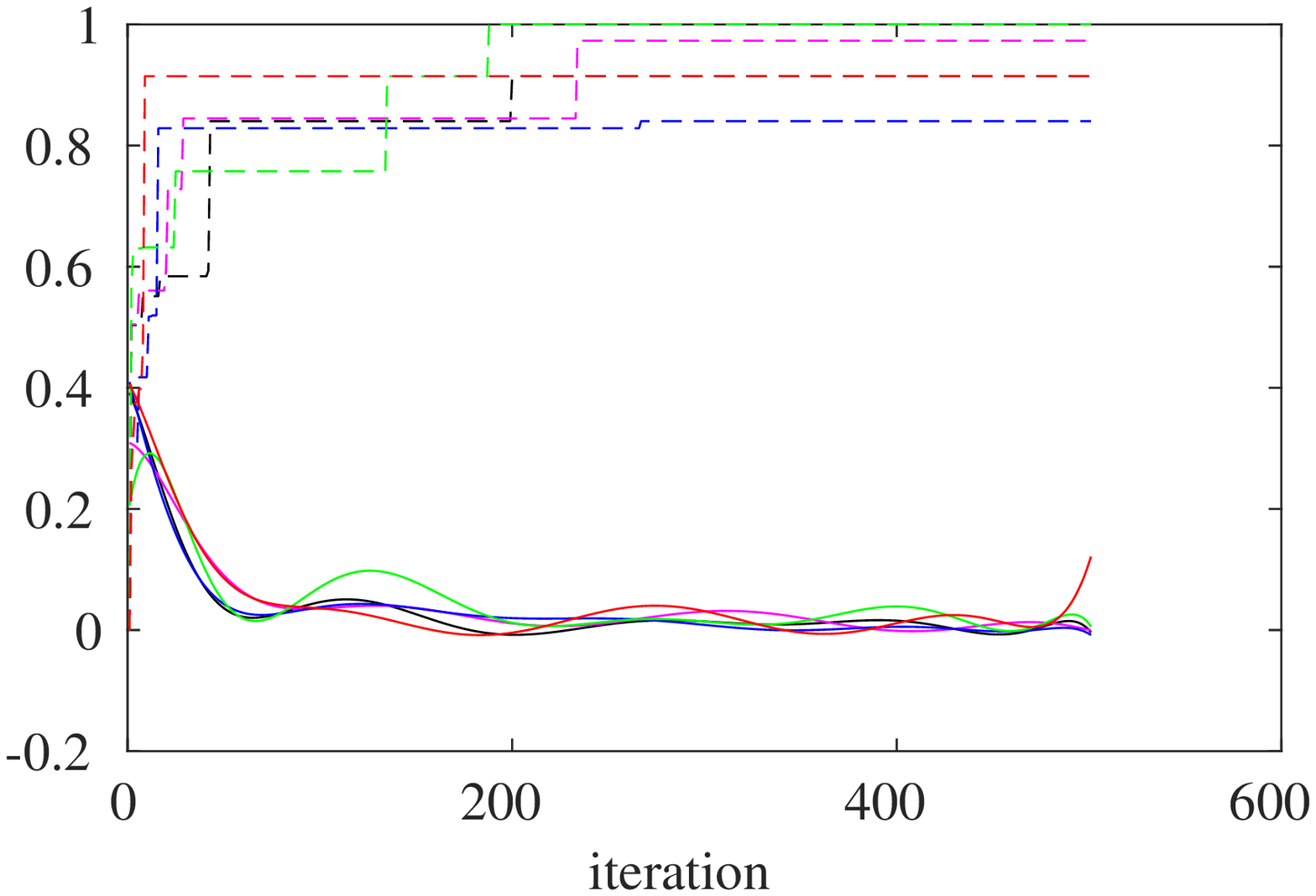}
  \label{fig:K10land5}
}
\quad 
\subfloat[K=12]{
  \centering
  \includegraphics[scale=0.26]{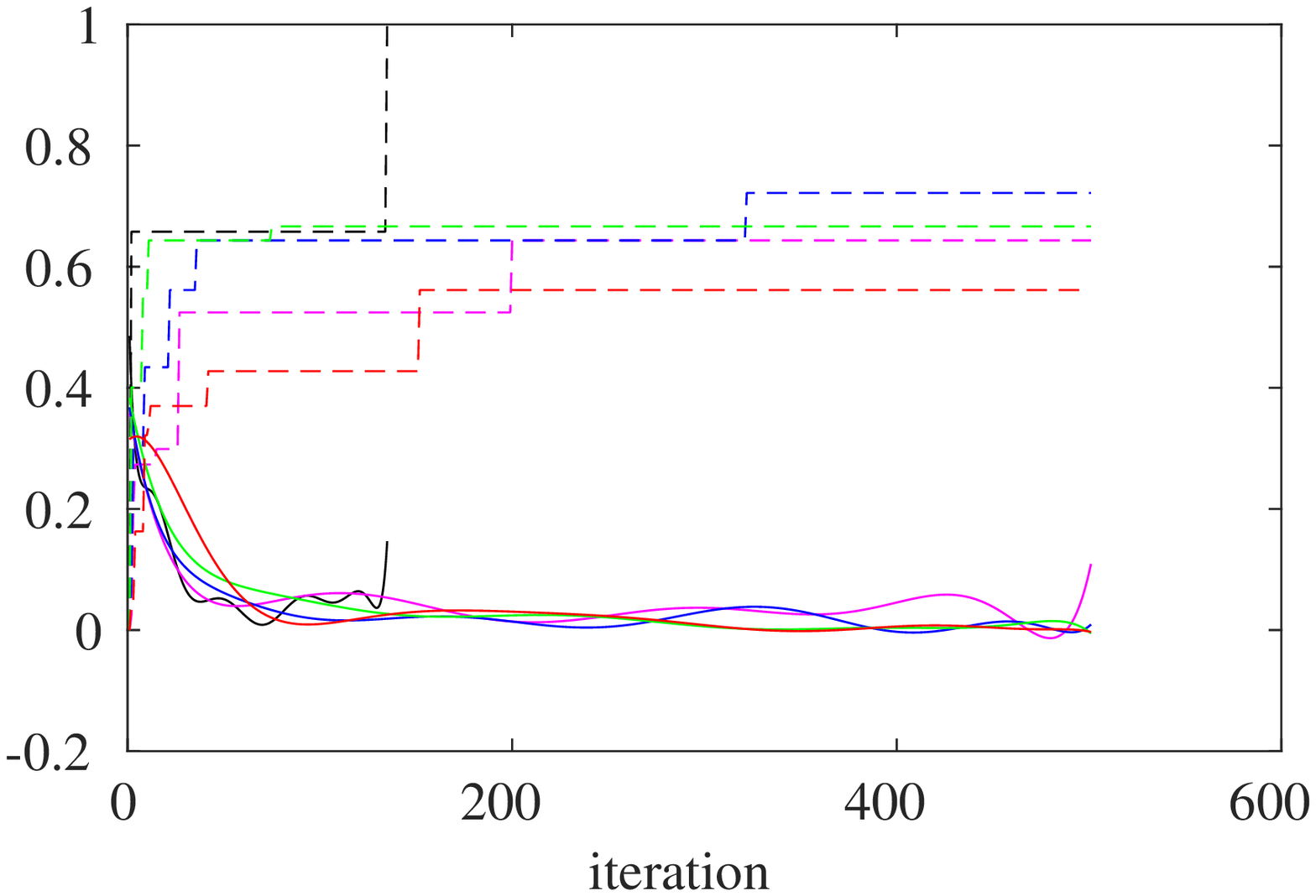}
  \label{fig:K12land5}
}
\quad 
\subfloat[K=14]{
  \centering
  \includegraphics[scale=0.26]{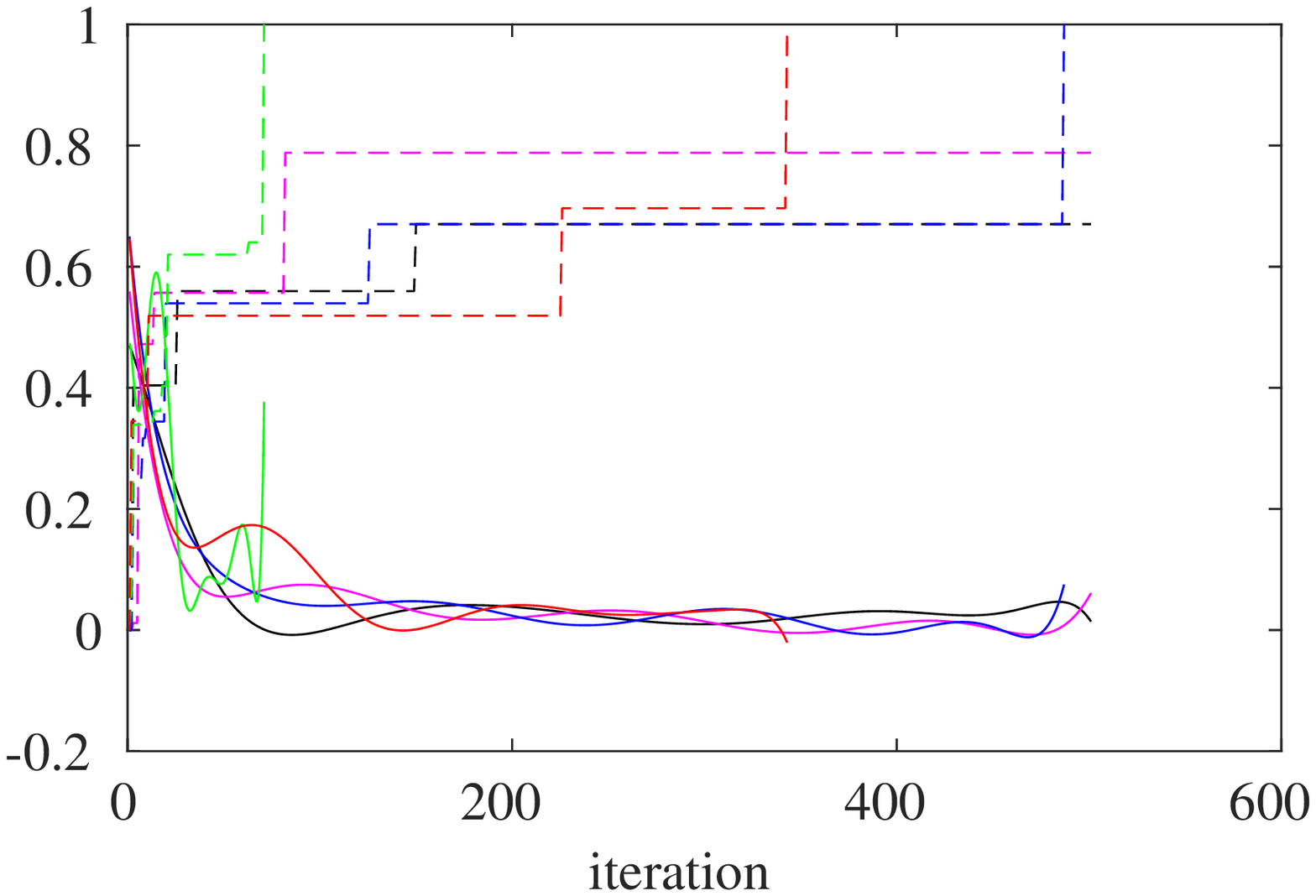}
  \label{fig:K14land5}
}
\quad 
\subfloat[K=16]{
  \centering
  \includegraphics[scale=0.26]{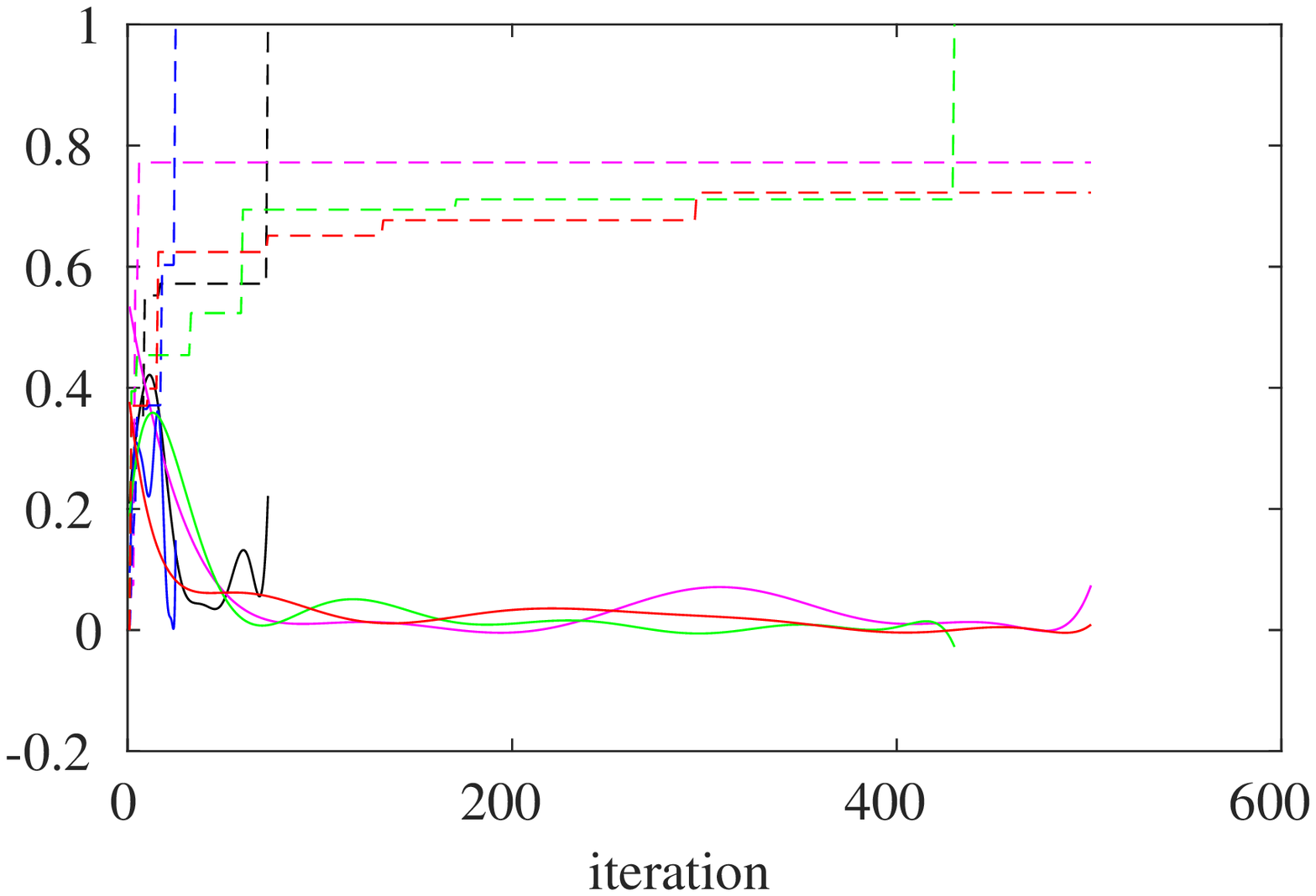}
  \label{fig:shd-fitness}
}
\caption{Illustration of the levels of similarities ({\it{SHD}}) in the generated Bayesian networks and the improvement in the objective values at each iteration. $N=18$.
\label{fig:shd-fitness}
}
\end{figure*}

The first observation we can make is that BOA converges faster for small values of $K$. The explanation of this behaviour is straightforward: these instances are easier to solve and therefore require less computational effort to solve. More importantly, we can see that significant improvements in the objective values do not necessarily correspond to significant changes in the PGM. This is our first indicator that adjusting the PGM at each iteration might not be the best strategy to adopt.

\begin{figure*}[h]
\centering
\subfloat[N=10]{
  \includegraphics[scale=.37]{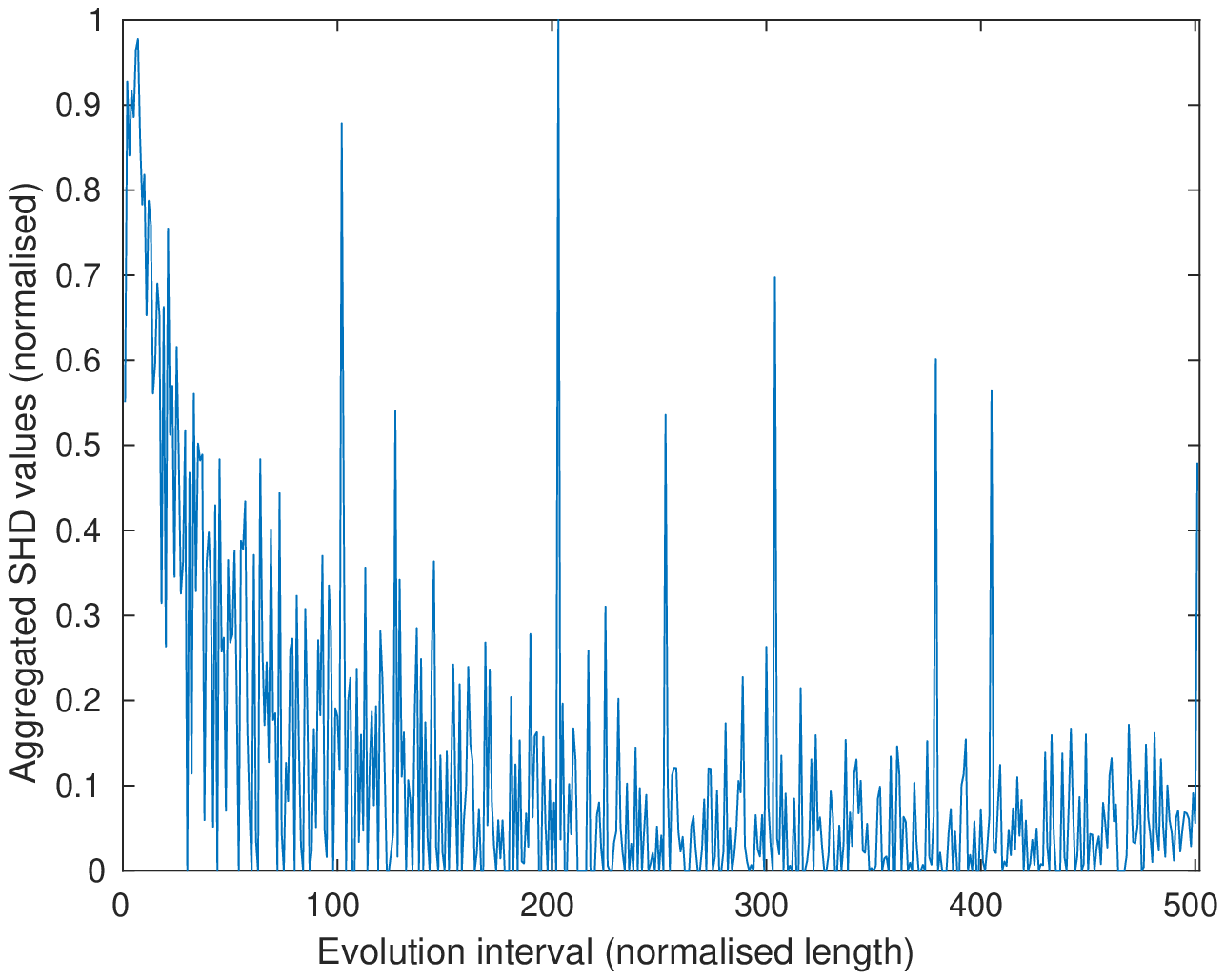}
}
\quad 
\subfloat[N=12]{
  \includegraphics[scale=.37]{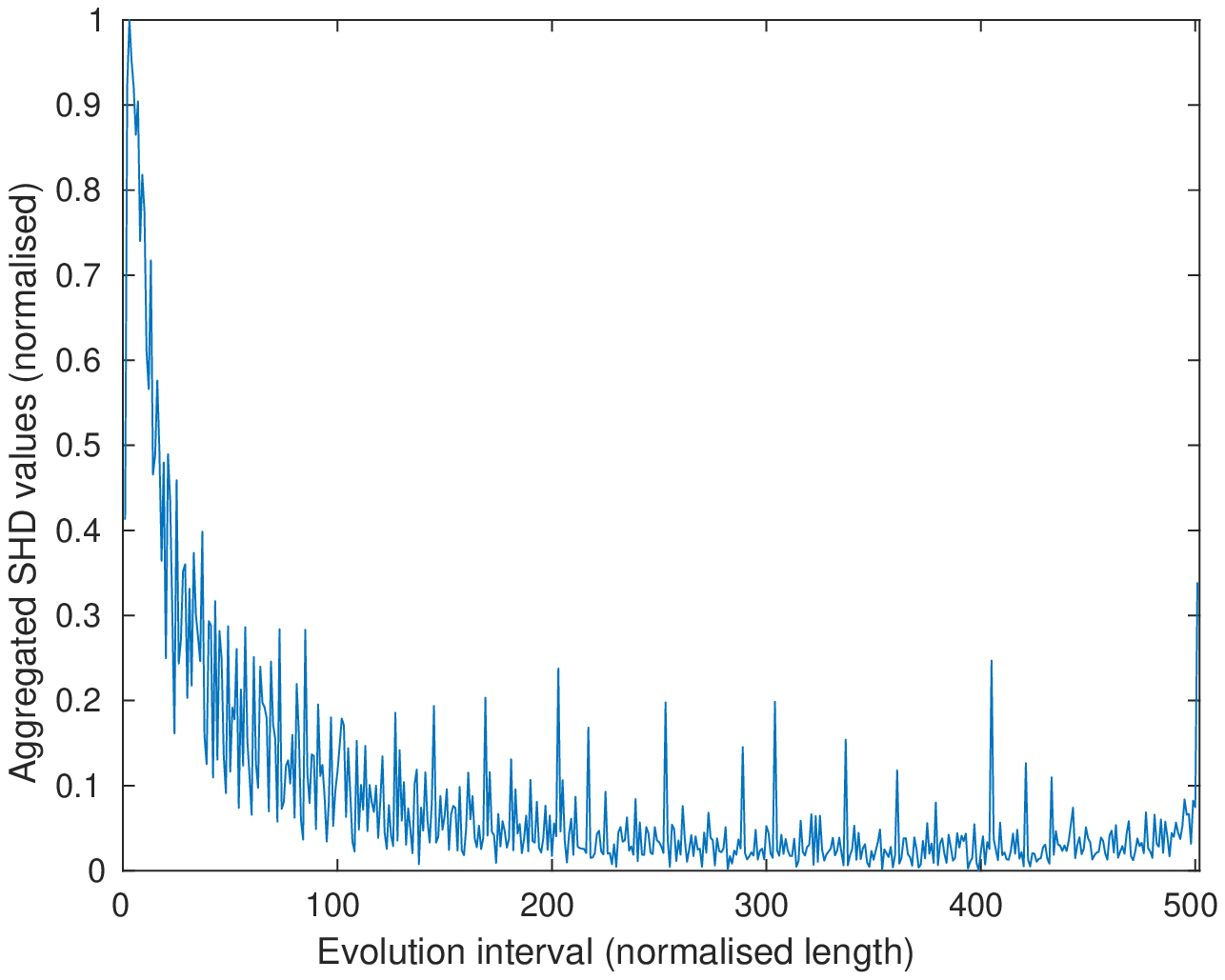}
}
\quad 
\subfloat[N=14]{
  \includegraphics[scale=.37]{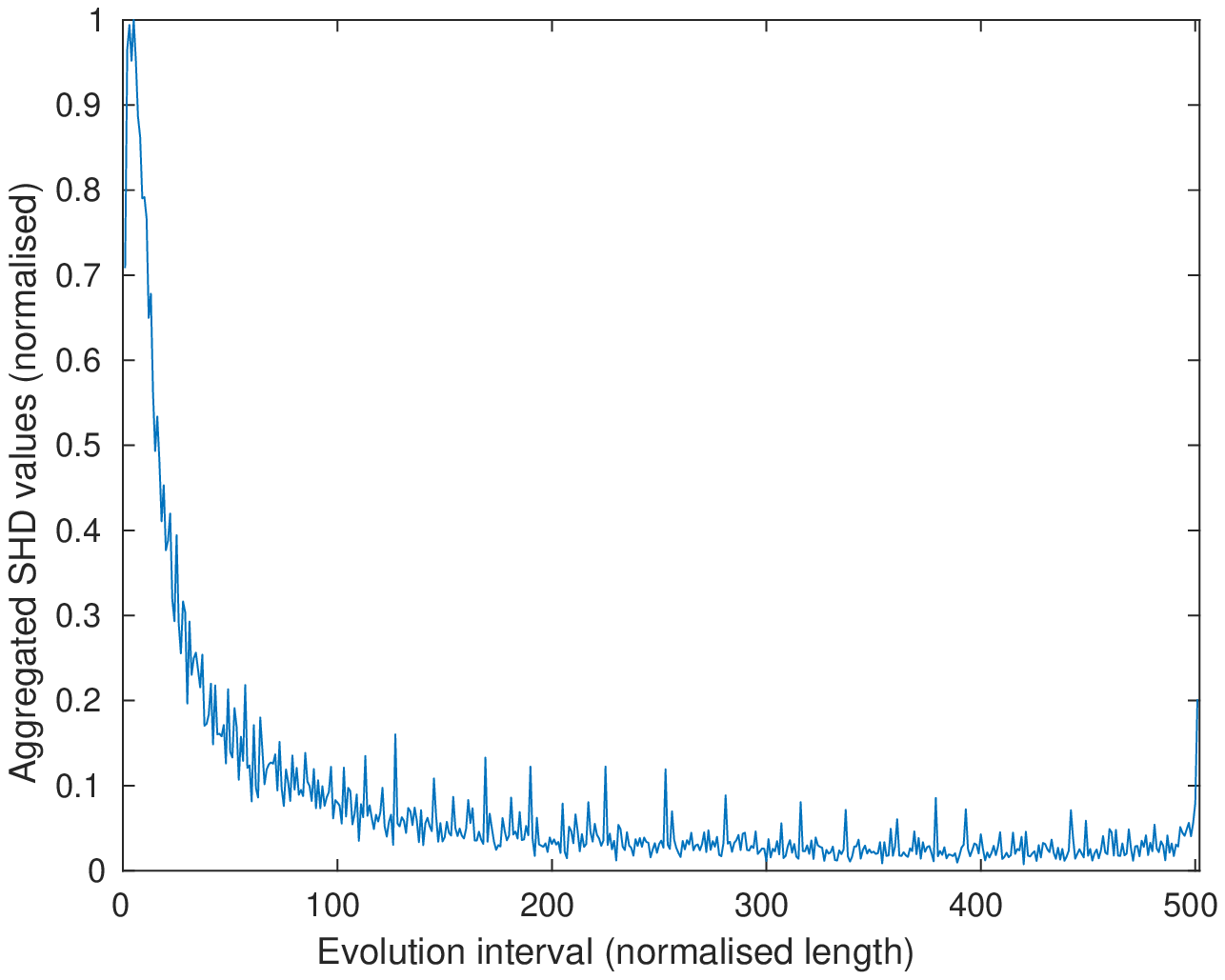}
}
\quad 
\subfloat[N=16]{
  \includegraphics[scale=.37]{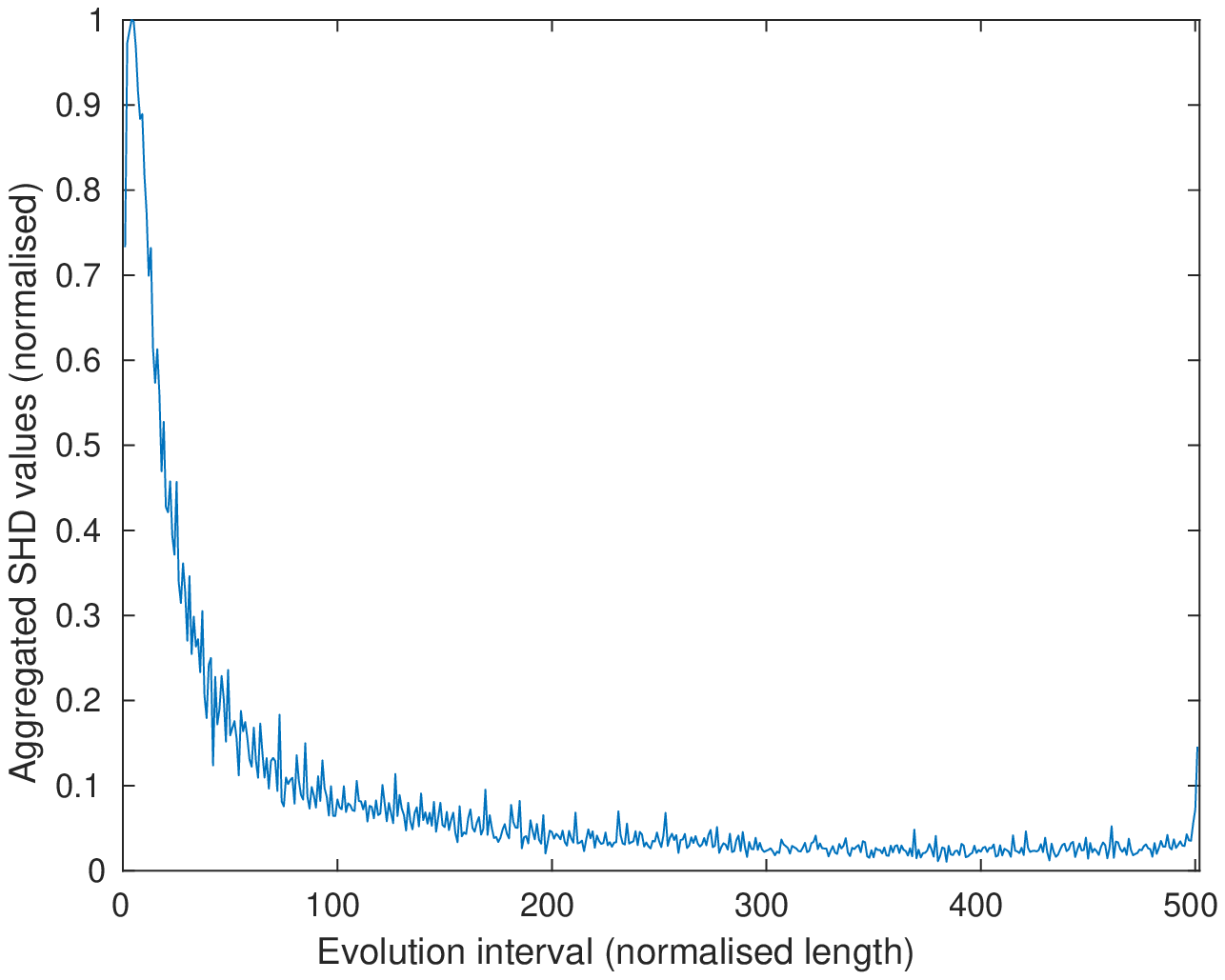}
}
\quad 
\subfloat[N=18]{
  \includegraphics[scale=.37]{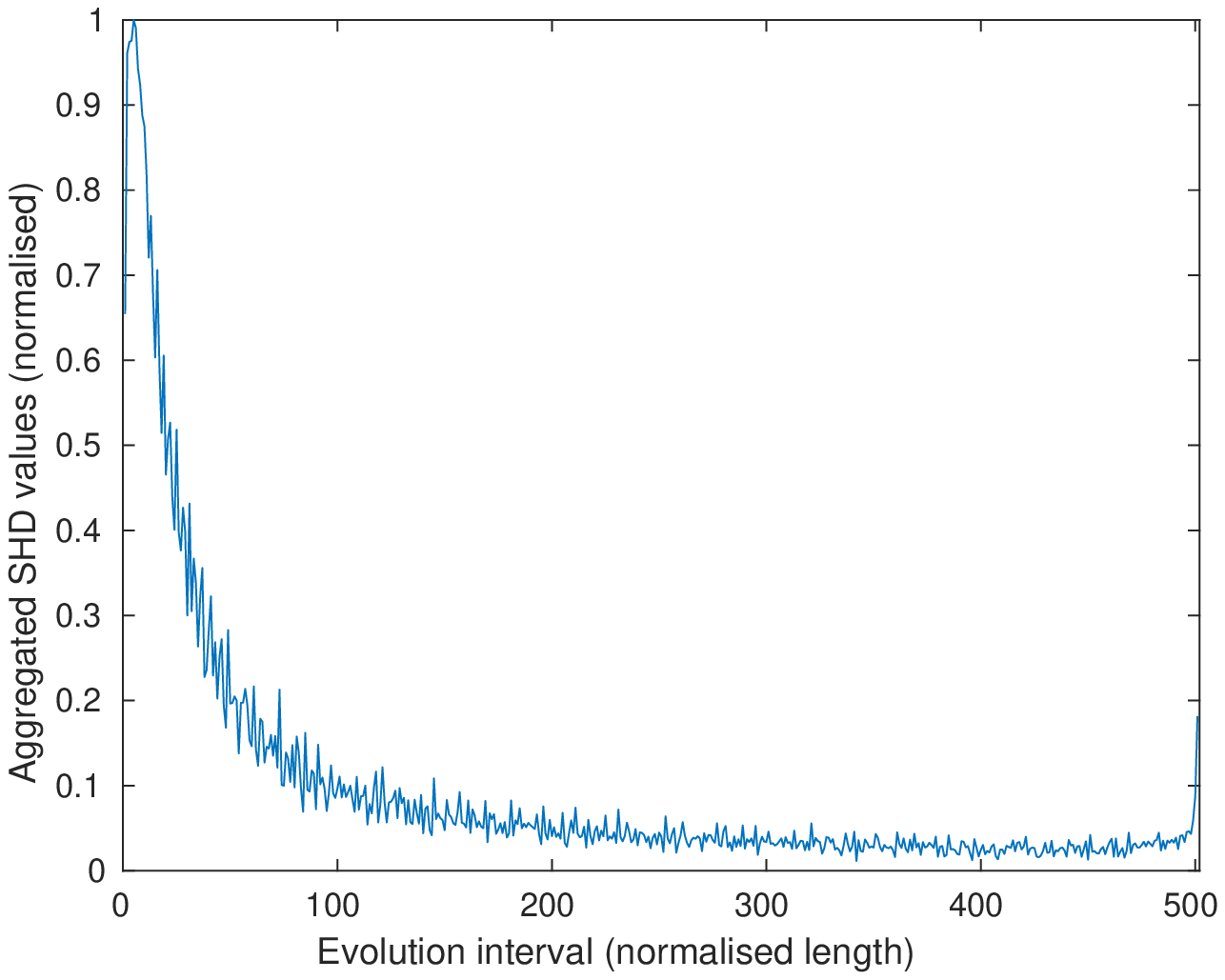}
}
\caption{
Bayesian network similarity patterns ($\overline{SHD}$) for different problem sizes. 
The curves mean values are 
(a) 0.125512;
(b) 0.091312;
(c) 0.083780;
(d) 0.088092;
(e) 0.097964.
\label{fig:shd-evolution}
}
\end{figure*}

Figure~\ref{fig:shd-evolution} represents the evolution of $\overline{SHD}$ for different problem sizes ($N=\{10, 12, 14, 16, 18\}$) and $L=500$. The number of iterations $L$ and values have both been normalised using the method proposed in Section \ref{sec:sect3}.

Looking at the patterns in Figure~\ref{fig:shd-evolution}, one would be tempted to use the Boltzmann criterion to decide whether to adjust the PGM or resample from the previous one. A good argument not to use the Boltzmann criterion is that the patterns in Figure~\ref{fig:shd-evolution} do not necessarily embed a Boltzmann distribution. In fact, we can identify a small, but significant, peak towards the end of the evolutionary process. This means that the latest generated PGMs are different from their predecessors. In other terms, the algorithm tends to diversify the search process before convergence by sampling from new probability distributions.

\subsection{The proposed algorithm}
\label{sec:FBOA}



Our proposed approach is summarised in Algorithm \ref{alg:boa2_framework}. It follows almost the same steps performed by BOA (Algorithm \ref{alg:boa_framework}), the exceptions are the use of a probability vector in line~\ref{boa2:probability-vector} and the decision rule in line~\ref{boa2:adjustment}.

In line \ref{boa2:probability-vector}, FBOA creates the PGM adjustment probability vector (which corresponds to the $\overline{SHD}$ vector for $N=18$ in our experimental study). 
The PGM adjustment at a given iteration $t$ is done with a probability $p_t$ as shown in line \ref{boa2:adjustment}. 
We will refer to this algorithm as FBOA, which stands for Fast Bayesian Optimisation Algorithm.


\begin{algorithm}[h]
\small\caption{FBOA algorithm} \label{alg:boa2_framework}
\begin{algorithmic}[1]
\STATE $t \gets 1$
\STATE $M^{(1)} \gets initial\_model()$ \label{boa2:generate-pgm}
\STATE ${p_t} \gets PGM\_probability\_vector()$ \label{boa2:probability-vector}
\REPEAT
  \STATE $S \gets sample\_solutions(M^{(t)})$ \label{step:sampling}
  \STATE $F \gets evaluate(S)$ \label{step:evaluation}
  \STATE $P \gets best\_solutions(S,F)$ \label{step:selection}
  \IF{$random(0,1) < p_t$} \label{boa2:adjustment}
    \STATE $M^{(t+1)} \gets adjust\_model(P)$
  \ELSE
    \STATE $M^{(t+1)} \gets M^{(t)}$
  \ENDIF
  \STATE $t \gets t+1$
\UNTIL{termination criterion met}
\end{algorithmic}
\end{algorithm}

\section{Experiments and Results}
\label{sec:sect5}


In this section, we are interested in finding out the ability of FBOA (algorithm presented in Section \ref{sec:FBOA}) to obtain similar performance in comparison with BOA. In particular, we investigate the estimated runtime of FBOA and BOA necessary to identify the optimal solution on particular NK-landscapes instances.

For both algorithms, we consider the population size $\lambda = |P| = 40$ and sample size $\mu = |S| = 100$.
The addressed NK-landscapes have the problem size $N \in \{10, 12, 14, 16, 18\}$ and the epistatic degree $K \in \{2, 4, 6, 8, 10, 12, 14, 16\}$ (as applicable).  
In order to enumerate the solution space exhaustively for each $N$, we used the largest value of $N = 18$ that can still be analysed with reasonable computational resources. A set of $10$ different landscapes are independently generated at random for each $N$ and $K$. 
Unless a near-optimal solution is found, a maximum number of evaluations $T_{\max} = 50000$ is used as a stopping criterion. Finally, each algorithm is executed $T_{\text{runs}}=100$ times per instance.

\subsection{Comparison of the optimisation results of BOA and FBOA}

First, in Table~\ref{tab:gap}, we report the average gaps between the optimal objective values (obtained through enumeration) and the best objective values found by BOA and FBOA on average for each combination of $K$ and $N$.
The p-values of the last column of Table~\ref{tab:gap} have been obtained using Friedman test \cite{siegel1988friedman} as the results are not normally distributed according to the Shapiro-Wilk normality test~\cite{conover:99}.
The Friedman test has been executed with a confidence level of $99\%$ indicating that there is statistically significant differences whenever $\text{p-value}<0.01$. According to Table~\ref{tab:gap}, there are no statistically significant differences between BOA and FBOA for all instances.  It means that BOA and FBOA performances cannot be differentiated based on the average gaps for all the considered combinations of $N$ and $K$.

\begin{table}[htbp]
\centering
\caption{
Average gap between the optimal and best objective values for BOA and FBOA
\label{tab:gap}
}
\begin{tabular}{c c c c c}
\toprule
\textbf{$N$}  &\textbf{$K$}    & \textbf{BOA}  & \textbf{FBOA}  &\textbf{p-value}  \\ \midrule
18	&$	2	$&$			0.0252		$&$		0.0203	$&$		0.8203	$\\
18	&$	4	$&$			0.0201		$&$		0.0193	$&$		0.8193	$\\
18	&$	6	$&$			0.0232		$&$		0.0245	$&$		0.8245	$\\
18	&$	8	$&$			0.0373		$&$		0.0364	$&$		0.8364	$\\
18	&$	10	$&$			0.0349		$&$		0.0348	$&$		0.8348	$\\
18	&$	12	$&$			0.0452		$&$		0.0443	$&$		0.8443	$\\
18	&$	14	$&$			0.0424		$&$		0.0406	$&$		0.8406	$\\
18	&$	16	$&$			0.0529		$&$		0.0511	$&$		0.8511	$\\
\midrule														
16	&$	2	$&$			0.0290		$&$		0.0316	$&$		0.8316	$\\
16	&$	4	$&$			0.0293		$&$		0.0320	$&$		0.8320	$\\
16	&$	6	$&$			0.0180		$&$		0.0191	$&$		0.8191	$\\
16	&$	8	$&$			0.0309		$&$		0.0312	$&$		0.8312	$\\
16	&$	10	$&$			0.0455		$&$		0.0452	$&$		0.8452	$\\
16	&$	12	$&$			0.0291		$&$		0.0296	$&$		0.8296	$\\
16	&$	14	$&$			0.0386		$&$		0.0366	$&$		0.8366	$\\
\midrule														
14	&$	2	$&$			0.0376		$&$		0.0342	$&$		0.8342	$\\
14	&$	4	$&$			0.0302		$&$		0.0319	$&$		0.8319	$\\
14	&$	6	$&$			0.0204		$&$		0.0213	$&$		0.8213	$\\
14	&$	8	$&$			0.0253		$&$		0.0250	$&$		0.8250	$\\
14	&$	10	$&$			0.0355		$&$		0.0347	$&$		0.8347	$\\
14	&$	12	$&$			0.0272		$&$		0.0287	$&$		0.8287	$\\
\midrule														
12	&$	2	$&$			0.0383		$&$		0.0381	$&$		0.8381	$\\
12	&$	4	$&$			0.0299		$&$		0.0303	$&$		0.8303	$\\
12	&$	6	$&$			0.0297		$&$		0.0245	$&$		0.8245	$\\
12	&$	8	$&$			0.0145		$&$		0.0145	$&$		0.8145	$\\
12	&$	10	$&$			0.0252		$&$		0.0255	$&$		0.8255	$\\
\midrule														
10	&$	2	$&$			0.0336		$&$		0.0319	$&$		0.8319	$\\
10	&$	4	$&$			0.0531		$&$		0.0488	$&$		0.8488	$\\
10	&$	6	$&$			0.0232		$&$		0.0196	$&$		0.8196	$\\
10	&$	8	$&$			0.0235		$&$		0.0231	$&$		0.8231	$\\
\bottomrule
\end{tabular}
\end{table}

Next, Table \ref{tab:success-rate} presents the average success rates for both BOA and FBOA with multiple $K$ and $N$ settings. The values represent averages of the number of successful runs over the total numbers of landscapes and runs.
The last row of each $N$ reports the average success rate over $K$. 

We apply the McNemar Chi-Square statistical test~\cite{mcnemar1947note} to compare BOA and FBOA approaches. The test is useful to show the difference between paired proportions and can determine whether there is marginal homogeneity between the two approaches.
There are a few statistically significant differences between success rates of two approaches when $\text{p-value}<0.01$ (confidence level of 99\%), which are highlighted. However, overall, the success rates are comparable. 

We can observe from Table~\ref{tab:success-rate} that there is no statistically significant difference for 26 of 30 NK-landscape configurations, except for those with $N=16$ ($K=8$ and $K=10$), and $N=18$ ($K=14$ and $K=16$) with some advantage for BOA. By considering the average over $K$ for each $N$ (last row of each $N$), there is no statistically significant difference between the considered approaches. It means that BOA and FBOA performances cannot be differentiated regarding the success rate for all the considered combinations of $N$ and $K$.

\begin{table}[htbp]
\centering
\caption{
Average success rates for BOA and FBOA. Four case of statistically significant differences are highlighted.
\label{tab:success-rate}
}
\begin{tabular}{c c c c c}
\toprule
\textbf{$N$} &\textbf{$K$} &  \textbf{BOA} &   \textbf{FBOA}  & \textbf{p-value} \\ \midrule
10	&$	2	$&$		1.0000		$&$		1.0000	$&$		0.9822	$\\
10	&$	4	$&$		0.9990		$&$		0.9990	$&$		0.9822	$\\
10	&$	6	$&$		0.9690		$&$		0.9740	$&$		0.9277	$\\
10	&$	8	$&$		0.9950		$&$		0.9950	$&$		0.9821	$\\
10	&	all	&$		0.9908		$&$		0.9920	$&$		0.9215	$\\
\midrule
12	&$	2	$&$		1.0000		$&$		1.0000	$&$		0.9822	$\\
12	&$	4	$&$		0.9820		$&$		0.9830	$&$		0.9987	$\\
12	&$	6	$&$		0.8980		$&$		0.9030	$&$		0.9249	$\\
12	&$	8	$&$		0.8570		$&$		0.8550	$&$		0.9807	$\\
12	&$	10	$&$		0.9280		$&$		0.9040	$&$		0.5910	$\\
12	&	all	&$		0.9330		$&$		0.9290	$&$		0.8548	$\\
\midrule
14	&$	2	$&$		0.9990		$&$		1.0000	$&$		0.9898	$\\
14	&$	4	$&$		0.9000		$&$		0.9300	$&$		0.4978	$\\
14	&$	6	$&$		0.7140		$&$		0.6630	$&$		0.1778	$\\
14	&$	8	$&$		0.5090		$&$		0.5350	$&$		0.4391	$\\
14	&$	10	$&$		0.4790		$&$		0.4570	$&$		0.4925	$\\
14	&$	12	$&$		0.4200		$&$		0.4870	$&$		0.0284	$\\
14	&	all	&$		0.6702		$&$		0.6787	$&$		0.6210	$\\
\midrule
16	&$	2	$&$		0.9950		$&$		0.9960	$&$		0.9899	$\\
16	&$	4	$&$		0.8620		$&$		0.8560	$&$		0.9040	$\\
16	&$	6	$&$		0.7090		$&$		0.7390	$&$		0.4460	$\\
16	&$	8	$&$		0.4300		$&$		0.2720	$&$	\textit{\textbf{0.0000}}	$\\
16	&$	10	$&$		0.3280		$&$		0.2600	$&$	\textit{\textbf{0.0057}}	$\\
16	&$	12	$&$		0.1960		$&$		0.1910	$&$		0.8389	$\\
16	&$	14	$&$		0.2330		$&$		0.2180	$&$		0.5097	$\\
16	&   all &$		0.5361		$&$		0.5046	$&$		0.0512	$\\
\midrule
18	&$	2	$&$		0.9720		$&$		0.9760	$&$		0.9458	$\\
18	&$	4	$&$		0.8340		$&$		0.7310	$&$		0.0100	$\\
18	&$	6	$&$		0.6440		$&$		0.5550	$&$		0.0110	$\\
18	&$	8	$&$		0.4260		$&$		0.4070	$&$		0.5328	$\\
18	&$	10	$&$		0.1610		$&$		0.1090	$&$		0.0191	$\\
18	&$	12	$&$		0.0480		$&$		0.0640	$&$		0.1564	$\\
18	&$	14	$&$		0.0800		$&$		0.0430	$&$	\textit{\textbf{0.0012}}	$\\
18	&$	16	$&$		0.1010		$&$		0.0570	$&$	\textit{\textbf{0.0006}}	$\\
18	&   all &$		0.4083		$&$		0.3678	$&$		0.0190	$\\
\bottomrule
\end{tabular}
\end{table}

\subsection{Comparison of the estimated runtimes of BOA and FBOA}

In this section, we are interested in comparing the estimated runtimes of BOA and FBOA for different values of $K$ (ruggedness).
A correlation analysis is performed between $K$ and the estimated runtime $ert$ given by Equation (\ref{eq:ert}), and a linear regression model is built according to Equation (\ref{eq:lin-reg})
\begin{equation}\label{eq:lin-reg}
ert=\beta_0+\beta_1.v_1+e
\end{equation}
using $ert$ as response variable, $v_1=K$ as explanatory variable, and a usual error $e$. A log-transformation is applied on both variables to better approximate linearity. The accuracy of the linear regression model is measured by the coefficient of determination $r^2$ which ranges from $0$ to $1$.

Table~\ref{tab:correlation-distance} shows the coefficient of determination for BOA and FBOA. The average distances between the respective linear regression curves are not log-transformed. According to Table~\ref{tab:correlation-distance}, well-adjusted linear regressions are obtained for all $N$ but $N=10$. 
For the best adjusted regression model $N=18$ (greatest coefficient of determination for each algorithm), FBOA is $3.8$ times faster than BOA.
Additionally, according to the average distance between the regression curves, FBOA is about $2.8$ times faster than BOA. 



\begin{table}[ht]
\center
\caption{
Coefficient of determination $r^2$ for BOA and FBOA, and average distance between linear regression curves
\label{tab:correlation-distance}
}
\begin{tabular}{cccc}
\toprule
\multicolumn{1}{c}{\multirow{2}{*}{\textbf{N}}} & \multicolumn{2}{c}{\textbf{$r^2$}}          & \multicolumn{1}{c}{\multirow{2}{*}{\textbf{Average distance}}} \\ \cline{2-3}
\multicolumn{1}{c}{}                            & \multicolumn{1}{c}{\textbf{BOA}} & \multicolumn{1}{c}{\textbf{FBOA}} & \multicolumn{1}{c}{} \\ \midrule
$10$ & $0.4650$ & $0.5963$ & $1.3998$ \\
$12$ & $0.7531$ & $0.6987$ & $2.5485$ \\
$14$ & $0.8643$ & $0.8243$ & $4.0443$ \\
$16$ & $0.8750$ & $0.8360$ & $3.6310$ \\
$18$ & $0.8949$ & $0.8940$ & $3.7509$ \\
\bottomrule
\end{tabular}
\end{table}

\begin{figure*}[ht]
\center
\subfloat[N=10]{
  \centering
  \includegraphics[scale=0.25]{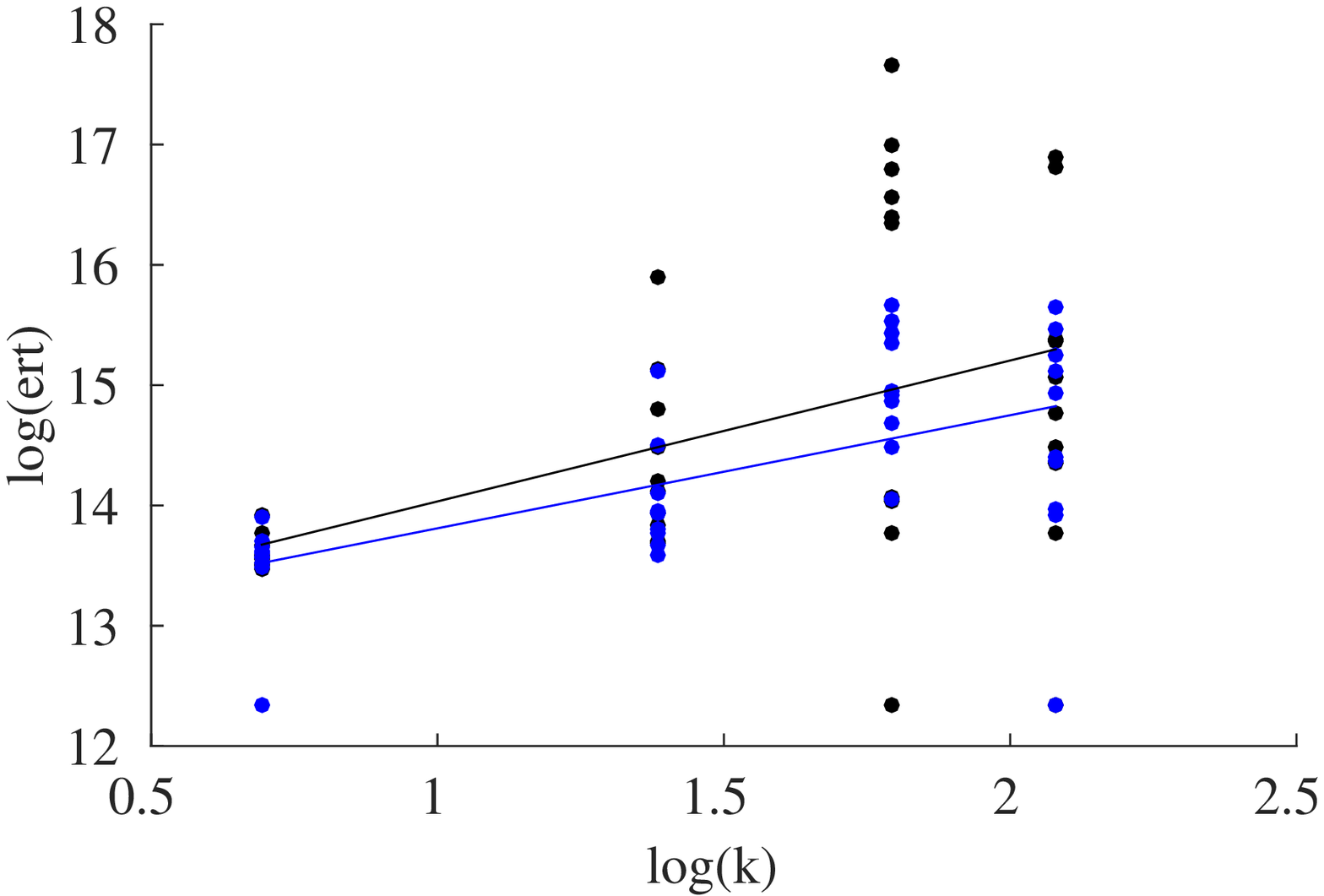}
}
\quad 
\subfloat[N=12]{
  \centering
  \includegraphics[scale=0.25]{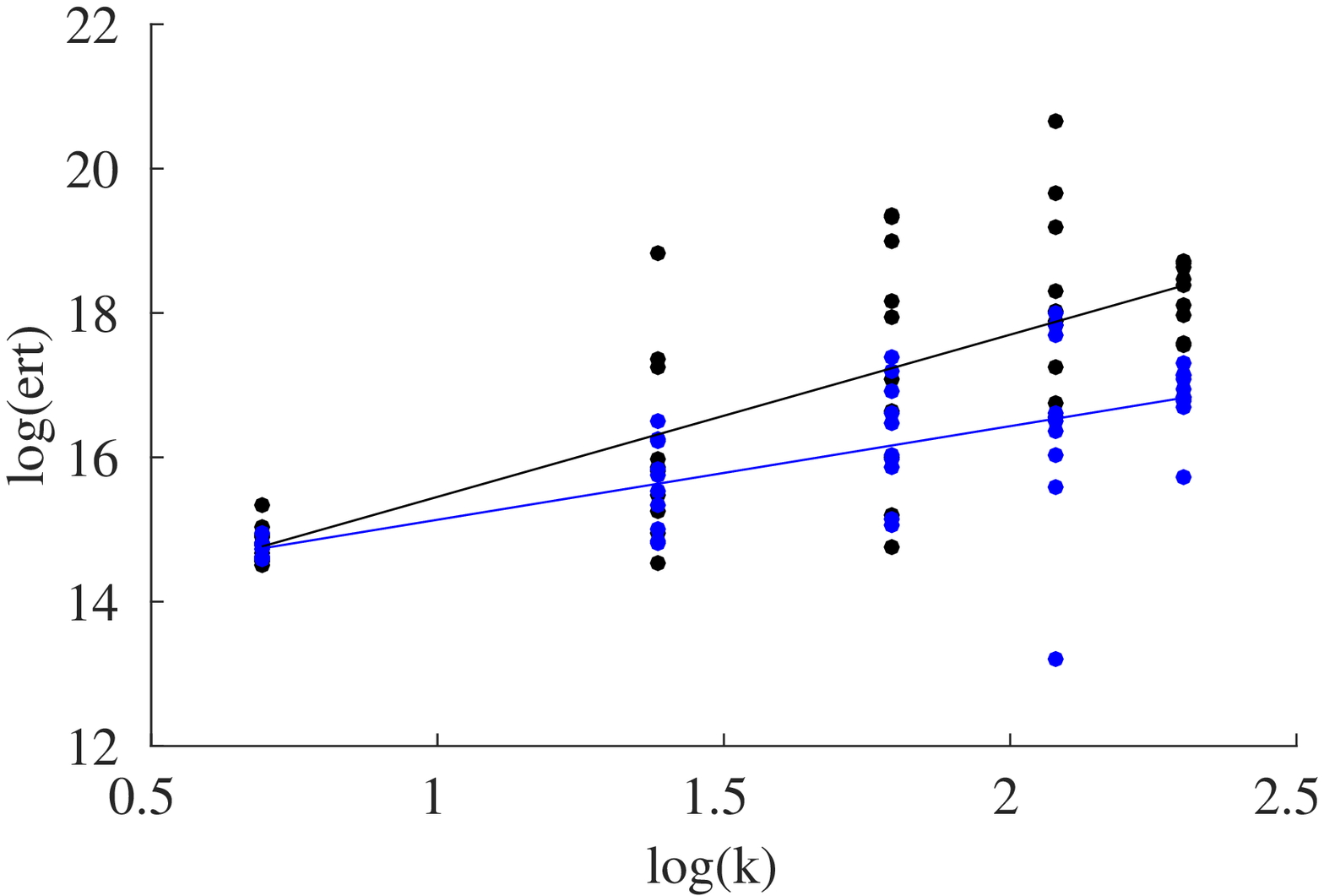}
}
\quad 
\subfloat[N=14]{
  \centering
  \includegraphics[scale=0.25]{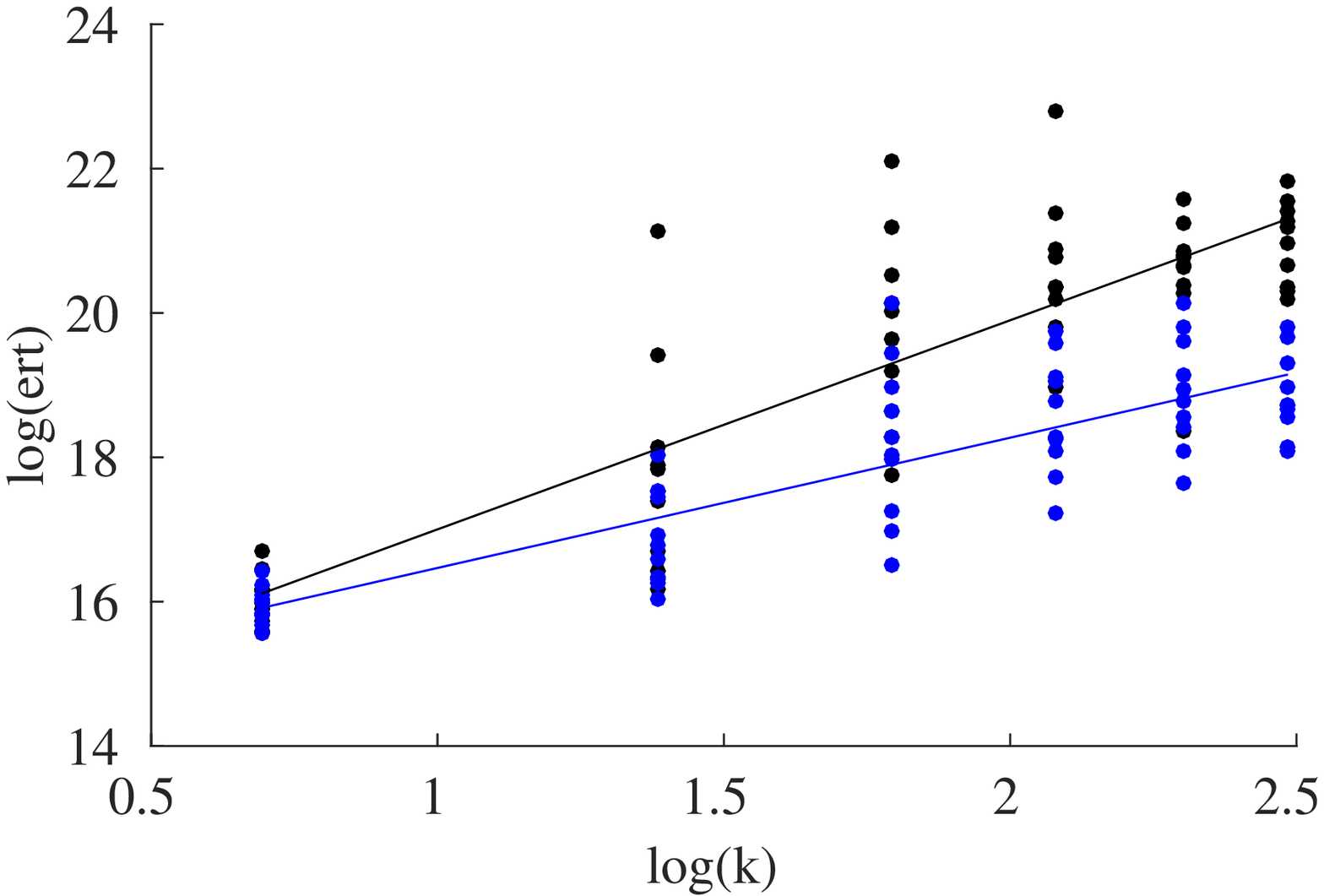}
}
\quad 
\subfloat[N=16]{
  \centering
  \includegraphics[scale=0.25]{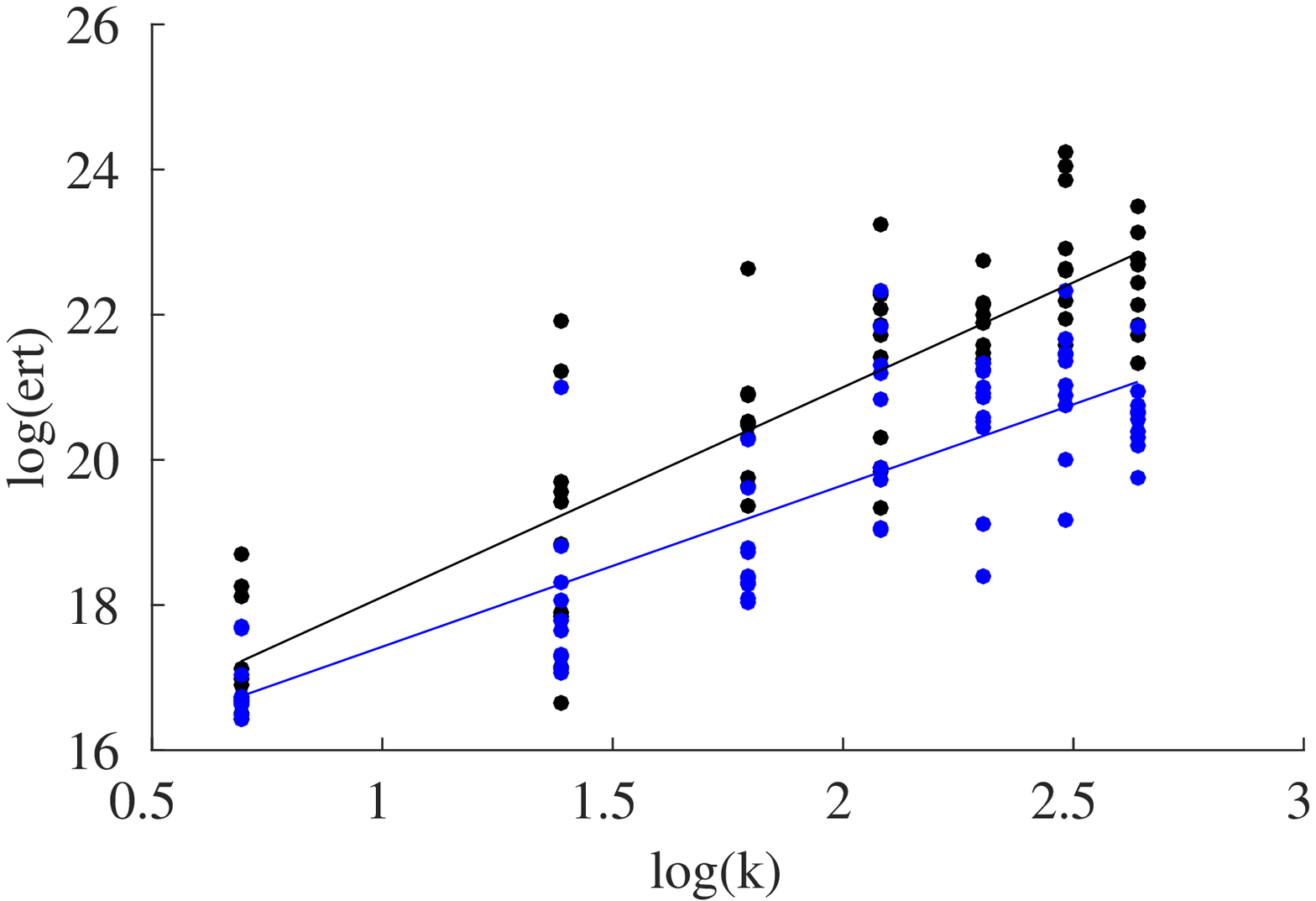}
}
\quad 
\subfloat[N=18]{
  \centering
  \includegraphics[scale=0.25]{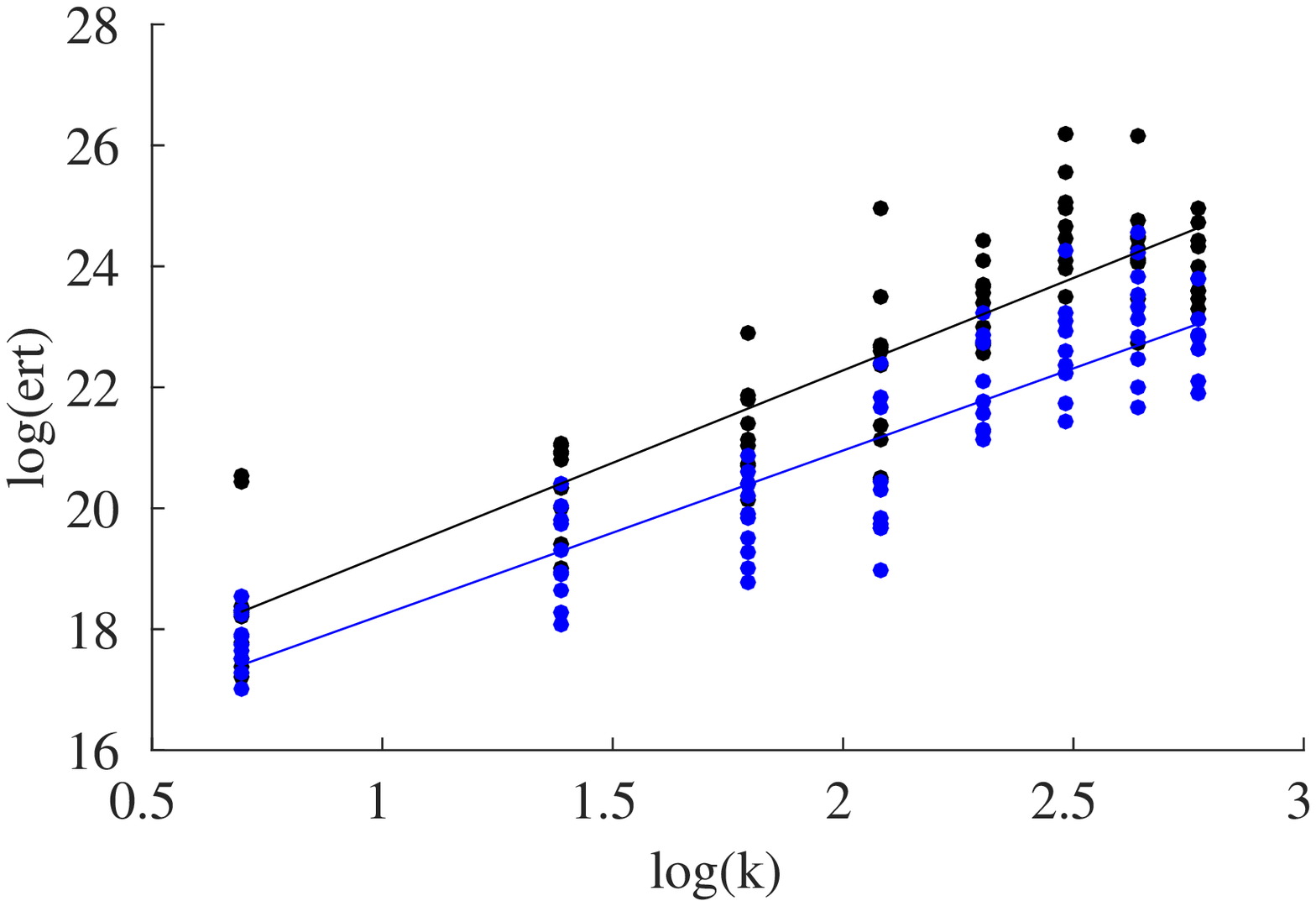}
}

\caption{
Correlation between $K$ and ert for BOA (black) and FBOA (blue) (in log-log scale)
\label{fig:regression-log}
}
\end{figure*}

Figure~\ref{fig:regression-log} shows both scatter plots and linear regression curves for all $N$ in a log-log scale. Each combination of $N$ and $K$ has $10$ values corresponding to ten different landscapes, and each landscape has its corresponding $ert$ given by Equation \ref{eq:ert}. According to the regression curves in Figure~\ref{fig:regression-log}, FBOA is always faster than BOA. It is worth noticing that runtime performances of FBOA increases for high values of $K$ as the slopes of regression curves of BOA are always smaller than those of BOA.

The results shown in Figure~\ref{fig:regression-log} and Table~\ref{tab:correlation-distance} might indicate that the computational effort saved with PGM adjustments could be beneficial as the problem difficulty increases (by increasing $N$ and $K$).
We do not claim that FBOA guarantees the best tradeoff between optimization results and runtime; this requires further investigation. 
However, the experiments show clearly that adjusting the PGM less often than what BOA suggests yields a significant runtime improvement without noticing any statistically significant differences between gaps obtained from BOA and FBOA to the optimal solutions.



\section{Conclusions and future directions} \label{sec:conclusion}

In this study, we investigated similarity patterns between two consecutive Bayesian networks using the Structural Hamming Distance (\emph{SHD}) metric, i.e. a proxy to measure a similarity between two networks, over the evolutionary process of BOA. The experiments clearly show patterns across a wide range of NK-landscapes in which the SHD values decrease during BOA's runs, implying the fact that generating a new Bayesian network at every iteration of the BOA might not be necessary -- except for an increase again towards the end that is necessary for exploitation.

Based on the this observation, we proposed a faster alternative called FBOA, which conducts the adjustment of the PGM following a probability as a function of the FBOA iteration. Furthermore, we tested the performance of FBOA in terms of its solution quality and computational burden. The experiments demonstrate that FBOA provides a solution quality comparable to BOA while dramatically saving computation time.

To further improve the performance of FBOA, one possibility is to incorporate the objective improvement information into the its decision of adjusting the PGM or resampling from the previous one. 
For example, we can employ self-adaptive methods that (1) enforce the generation of a new PGM or (2) increase the probability of generating a new PGM, if the best found objective value is not improved by sampling from the current PGM. 
Alternatively, the probability of generating a new PGM can be represented as a function of the iteration number and the improvements in the objective value. At a given iteration, the likelihood of resampling instead generating a new PGM is increased when the objective value is improved.


Although we have proposed an alternative implementation of BOA, this work is mainly to show that the frequency of adjusting the PGM can be tuned to save computational time. The subject requires more investigation in the future from both theoretical and empirical perspectives.

\bibliographystyle{unsrt}  
\bibliography{references}  

\end{document}